\def\year{2018}\relax
\documentclass[letterpaper]{article} 
\usepackage{aaai18}  
\usepackage{times}  
\usepackage{helvet}  
\usepackage{courier}  
\usepackage{url}  
\usepackage{color}
\usepackage{graphicx}  
\usepackage{amsmath}
\usepackage{amssymb}
\usepackage{subfigure}
\usepackage{epstopdf}
\usepackage{array}
\usepackage[]{booktabs}
\usepackage{multirow}
\usepackage{dsfont}

\allowdisplaybreaks
\makeatletter
\newcommand{\thickhline}{%
    \noalign {\ifnum 0=`}\fi \hrule height 1pt
    \futurelet \reserved@a \@xhline
}
\makeatother
\begin{document}

\title{Learning Pose Grammar to Encode Human Body Configuration for\\ 3D Pose Estimation}
\author{Hao-Shu Fang$^{1,2}$\thanks{Hao-Shu Fang, Yuanlu Xu and Wenguan Wang contributed equally to this paper. This work is supported by ONR MURI Project N00014-16-1-2007, DARPA XAI Award N66001-17-2-4029, and NSF IIS 1423305, 1657600. Hao-Shu Fang and Wenguan Wang are visiting students. The correspondence author is Xiaobai Liu.},\, Yuanlu Xu$^{1*}$,\, Wenguan Wang$^{1,3*}$, Xiaobai Liu$^4$,\, Song-Chun Zhu$^1$\\
$^1$Dept. Computer Science and Statistics, University of California, Los Angeles\;\;\; $^2$Shanghai Jiao Tong University\\
$^3$Beijing Institute of Technology\;\;\; $^4$Dept. Computer Science, San Diego State University\\
{\tt\small fhaoshu@gmail.com, yuanluxu@cs.ucla.edu, wenguanwang@bit.edu.cn}\\
{\tt\small xiaobai.liu@mail.sdsu.edu, sczhu@stat.ucla.edu}
}

\maketitle

\begin{abstract}
In this paper, we propose a pose grammar to tackle the problem of 3D human pose estimation. Our model directly takes 2D pose as input and learns a generalized 2D-3D mapping function. The proposed model consists of a base network which efficiently captures pose-aligned features and a hierarchy of Bi-directional RNNs (BRNN) on the top to explicitly incorporate a set of knowledge regarding human body configuration (\textit{i.e.}, kinematics, symmetry, motor coordination). The proposed model thus enforces high-level constraints over human poses. In learning, we develop a pose sample simulator to augment training samples in virtual camera views, which further improves our model generalizability. We validate our method on public 3D human pose benchmarks and propose a new evaluation protocol working on cross-view setting to verify the generalization capability of different methods. We empirically observe that most state-of-the-art methods encounter difficulty under such setting while our method can well handle such challenges.
\end{abstract}

\maketitle

\section{Introduction} \label{sec:intro}

Estimating 3D human poses from a single-view RGB image has attracted growing interest in the past few years for its wide applications in robotics, autonomous vehicles, intelligent drones etc.
This is a challenging inverse task since it aims to reconstruct 3D spaces from 2D data and the inherent ambiguity is further amplified by other factors, \textit{e.g.}, clothes, occlusions, background clutters. With the availability of large-scale pose datasets, \textit{e.g.}, Human3.6M~\cite{ionescu2014human3}, deep learning based methods have obtained encouraging success. These methods can be roughly divided into two categories: i) learning end-to-end networks that recover 2D input images to 3D poses directly, ii) extracting 2D human poses from input images and then lifting 2D poses to 3D spaces.

\begin{figure}[ptb]
\centering
\includegraphics[width=\linewidth]{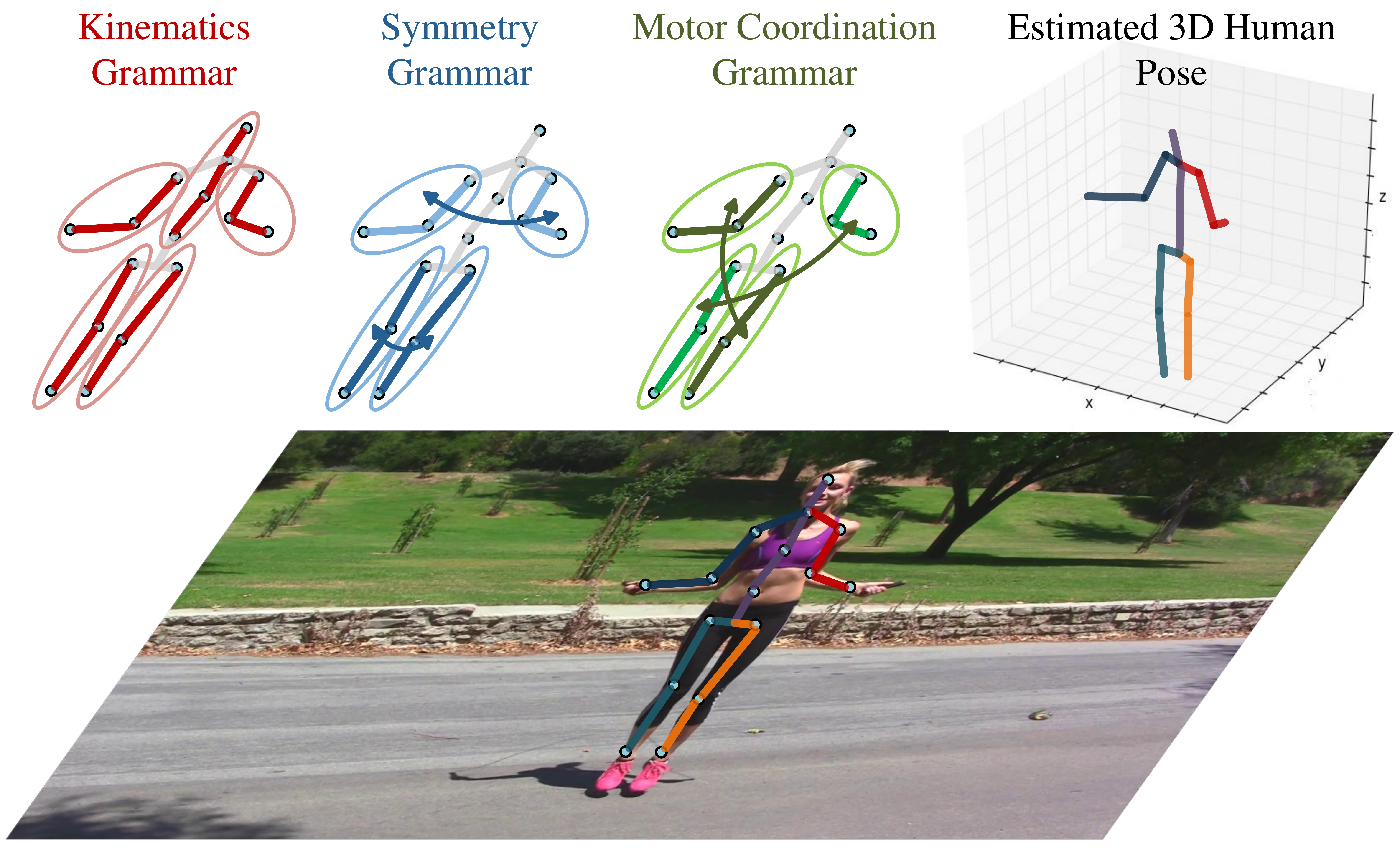}
\caption{Illustration of human pose grammar, which express the knowledge of human body configuration. We consider three kinds of human body dependencies and relations in this paper, \textit{i.e.}, kinematics (red), symmetry (blue) and motor coordination (green).}
\label{fig:intro}
\end{figure}

There are some advantages to decouple 3D human pose estimation into two stages. i) For 2D pose estimation, existing large-scale pose estimation datasets~\cite{andriluka20142d,PoseDataset16} have provided sufficient annotations; whereas pre-trained 2D pose estimators~\cite{newell2016stacked} are also generalized and mature enough to be deployed elsewhere. ii) For 2D to 3D reconstruction, infinite 2D-3D pose pairs can be generated by projecting each 3D pose into 2D poses under different camera views. Recent works~\cite{yasin2016dual,martinez2017simple} have shown that well-designed deep networks can achieve state-of-the-art performance on Human3.6M dataset using only 2D pose detections as system inputs.

However, despite their promising results, few previous methods explored the problem of encoding domain-specific knowledge into current deep learning based detectors.

In this paper, we develop a deep grammar network to explicitly encode a set of knowledge over human body dependencies and relations, as illustrated in Figure~\ref{fig:intro}. These knowledges explicitly express the composition process of joint-part-pose, including kinematics, symmetry and motor coordination, and serve as knowledge bases for reconstructing 3D poses. We ground these knowledges in a multi-level RNN network which can be end-to-end trained with back-propagation. The composed hierarchical structure describes composition, context and high-order relations among human body parts.

Additionally, we empirically find that previous methods are restricted to their poor generalization capabilities while performing cross-view pose estimation, \textit{i.e.}, being tested on human images from unseen camera views. Notably, on the Human3.6M dataset, the largest publicly available human pose benchmark, we find that the performance of state-of-the-art methods heavily relies on the camera viewpoints. As shown in Table 1, once we change the split of training and testing set, using 3 cameras for training and testing on the forth camera (new \textit{protocol \#3}), performance of state-of-the-art methods drops dramatically and is much worse than image-based deep learning methods. These empirical studies suggested that existing methods might over-fit to sparse camera settings and bear poor generalization capabilities.

To handle the issue, we propose to augment the learning process with more camera views, which explore a generalized mapping from 2D spaces to 3D spaces. More specifically, we develop a pose simulator to augment training samples with virtual camera views, which can further improve system robustness. Our method is motivated by the previous works on learning by synthesis. Differently, we focus on the sampling of 2D pose instance from a given 3D space, following the basic geometry principles. In particular, we develop a pose simulator to effectively generate training samples from unseen camera views. These samples can greatly reduce the risk of over-fitting and thus improve generalization capabilities of the developed pose estimation system.

We conduct exhaustive experiments on public human pose benchmarks, \textit{e.g.}, Human3.6M, HumanEva, MPII, to verify the generalization issues of existing methods, and evaluate the proposed method for cross-view human pose estimation. Results show that our method can significantly reduce pose estimation errors and outperform the alternative methods to a large extend.

\textbf{Contributions}. There are two major contributions of the proposed framework: i) a deep grammar network that incorporates both powerful encoding capabilities of deep neural networks and high-level dependencies and relations of human body; ii) a data augmentation technique that improves generalization ability of current 2-step methods, allowing it to catch up with or even outperforms end-to-end image-based competitors.

\section{Related Work} \label{sec:literature}

The proposed method is closely related to the following two tracks in computer vision and artificial intelligence.

\textbf{3D pose estimation}. In literature, methods solving this task can be roughly classified into two frameworks: i) directly learning 3D pose structures from 2D images, ii) a cascaded framework of first performing 2D pose estimation and then reconstructing 3D pose from the estimated 2D joints. Specifically, for the first framework,~\cite{li20143d} proposed a multi-task convolutional network that simultaneously learns pose regression and part detection.~\cite{tekin2016structured} first learned an auto-encoder that describes 3D pose in high dimensional space then mapped the input image to that space using CNN.~\cite{pavlakos2017volumetric} represented 3D joints as points in a discretized 3D space and proposed a coarse-to-fine approach for iterative refinement.~\cite{zhou2017towards} mixed 2D and 3D data and trained an unified network with two-stage cascaded structure. These methods heavily relies on well-labeled image and 3D ground-truth pairs, since they need to learn depth information from images.

To avoid this limitation, some work~\cite{paul2003fast,jiang20103d,yasin2016dual} tried to address this problem in a two step manner. For example, in~\cite{yasin2016dual}, the authors proposed an exemplar-based method to retrieve the nearest 3D pose in the 3D pose library using the estimated 2D pose. Recently,~\cite{martinez2017simple} proposed a network that directly regresses 3D keypoints from 2D joint detections and achieves state-of-the-art performance. Our work takes a further step towards a unified 2D-to-3D reconstruction network that integrates the learning power of deep learning and the domain-specific knowledge represented by hierarchy grammar model. The proposed method would offer a deep insight into the rationale behind this problem.

\textbf{Grammar model}. This track receives long-lasting endorsement due to its interpretability and effectiveness in modeling diverse tasks~\cite{LiuSIGGRAPH2014,xu2016multi,xu2017multi}. In~\cite{HanTPAMI2009}, the authors approached the problem of image parsing using a stochastic grammar model.  After that, grammar models have been used in~\cite{xu2013reid,xu2014search} for 2D human body parsing. \cite{ParkICCV15} proposed a phrase structure, dependency and attribute grammar for 2D human body, representing decomposition and articulation of body parts. Notably,~\cite{Nie3DPoseCVPR17} represented human body as a set of simplified kinematic grammar and learn their relations with LSTM. In this paper, our representation can be analogized as a hierarchical attributed grammar model, with similar hierarchical structures, BRNNS as probabilistic grammar. The difference lies in that our model is fully recursive and without semantics in middle levels.

\begin{figure*}[ptb]
\centering
\includegraphics[width=\linewidth]{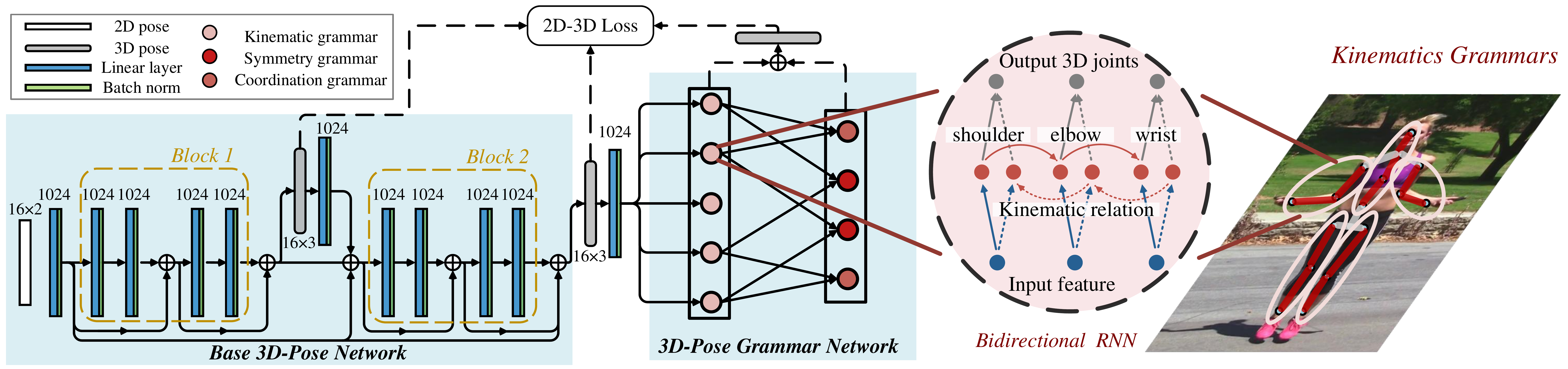}
\caption{The proposed deep grammar network. Our model consists of two major components: a base network constituted by two basic blocks and a pose grammar network encoding human body dependencies and relations w.r.t. kinematics, symmetry and motor coordination. Each grammar is represented as a Bi-directional RNN among certain joints. See text for detailed explanations.}
\label{fig:network}
\end{figure*}

\section{Representation} \label{sec:rep}

We represent the 2D human pose $\mathbf{U}$ as a set of $N_U$ joint locations
\begin{equation}
	\mathbf{U} = \{u_i:\, i = 1,\dots,N_U,\; u_i \in \mathbb{R}^2\}.
\end{equation}
Our task is to estimate the corresponding 3D human pose $\mathbf{V}$ in the world reference frame. Suppose the 2D coordinate of a joint $u_i$ is $[x_i, y_i]$ and the 3D coordinate $v_i$ is $[X_i, Y_i, Z_i]$, we can describe the relation between 2D and 3D as a pinhole image projection
\begin{equation}\small\begin{aligned}
    \begin{bmatrix} x_i\\ y_i\\ w_i \end{bmatrix}
    = K \left[R|RT\right] \begin{bmatrix} X_i\\ Y_i\\ Z_i \\ 1 \end{bmatrix},
    K = \begin{bmatrix} \alpha_x &0 &x_0\\ 0 &\alpha_y &y_0\\ 0 &0 &1 \end{bmatrix},
    T = \begin{bmatrix} T_x \\ T_y\\ T_z \end{bmatrix},
\end{aligned}\end{equation}
where $w_i$ is the depth w.r.t. the camera reference frame, $K$ is the camera intrinsic parameter (\textit{e.g.}, focal length $\alpha_x$ and $\alpha_y$, principal point $x_0$ and $y_0$), $R$ and $T$ are camera extrinsic parameters of rotation and translation, respectively. Note we omit camera distortion for simplicity.

It involves two sub-problems in estimating 3D pose from 2D pose: i) calibrating camera parameters, and ii) estimating 3D human joint positions. Noticing that these two sub-problems are entangled and cannot be solved without ambiguity, we propose a deep neural network to learn the generalized 2D$\rightarrow$3D mapping $\mathbf{V} = f(\mathbf{U};\theta)$, where $f(\cdot)$ is a multi-to-multi mapping function, parameterized by $\theta$.

\subsection{Model Overview}

Our model follows the line that directly estimating 3D human keypoints from 2D joint detections, which renders our model high applicability. More specifically, we extend various human pose grammar into deep neural network, where a basic 3D pose detection network is first used for extracting pose-aligned features, and a hierarchy of RNNs is built for encoding high-level 3D pose grammar for generating final reasonable 3D pose estimations. Above two networks work in a cascaded way, resulting in a strong 3D pose estimator that inherits the representation power of neural network and high-level knowledge of human body configuration.

\subsection{Base 3D-Pose Network}

For building a solid foundation for high-level grammar model, we first use a base network for capturing well both 2D and 3D pose-aligned features. The base network is inspired by \cite{martinez2017simple}, which has been demonstrated effective in encoding the information of 2D and 3D poses. As illustrated in Figure~\ref{fig:network}, our base network consists of two cascaded blocks. For each block, several linear (fully connected) layers, interleaved with Batch Normalization, Dropout layers, and \textit{ReLU} activation, are stacked for efficiently mapping the 2D-pose features to higher-dimensions.
The input 2D pose detections $\mathbf{U}$ (obtained as ground truth 2D joint locations under known camera parameters, or from other 2D pose detectors) are first projected into a $1024$-\textit{d} features, with a fully connected layer. Then the first block takes this high-dimensional features as input and an extra linear layer is applied at the end of it to obtain an explicit 3D pose representation. In order to have a coherent understanding of the full body in 3D space, we re-project the 3D estimation into a $1024$-dimension space and further feed it into the second block. With the initial 3D pose estimation from the first block, the second block is able to reconstruct a more reasonable 3D pose.
To take a full use of the information of initial 2D pose detections, we introduce \textit{residual connections}~\cite{he2016deep} between the two blocks. Such technique is able to encourage the information flow and facilitate our training. Additionally, each block in our base network is able to directly access to the gradients from the loss function (detailed in Sec.\ref{sec:learn}), leading to an implicit deep supervision~\cite{lee2015deeply}.
With the refined 3D-pose, estimated from base network, we again re-projected it into a $1024$-\textit{d} features. We combine the $1024$-\textit{d} features from the 3D-pose and the original $1024$-\textit{d} feature of 2D-pose together, which leads to a powerful representation that has well-aligned 3D-pose information and preserves the original 2D-pose information. Then we feed this feature into our 3D-pose grammar network.

\subsection{3D-Pose Grammar Network}

So far, our base network directly estimated the depth of each joint from the 2D pose detections. However, the natural of human body that rich inherent structures are involved in this task, motivates us to reason the 3D structure of the whole person in a global manner. Here we extend Bi-directional RNNs (BRNN) to model high-level knowledge of 3D human pose grammar, which towards a more reasonable and powerful 3D pose estimator that is capable of satisfying human anatomical and anthropomorphic constraints. Before going deep into our grammar network, we first detail our grammar formulations that reflect interpretable and high-level knowledge of human body configuration. Basically, given a human body, we consider the following three types of grammar in our network.

\textbf{Kinematic grammar} $\mathcal{G}^{kin}$ describes human body movements without considering forces (\textit{i.e.}, the red skeleton in Figure~\ref{fig:intro})). We define 5 kinematic grammar to represent the constraints among kinematically connected joints:
\begin{align}
    \mathcal{G}^{kin}_{spine}&:\; \textit{head} \leftrightarrow \textit{thorax} \leftrightarrow \textit{spine} \leftrightarrow \textit{hip}\;, \\
    \mathcal{G}^{kin}_{l.arm}&:\; \textit{l.shoulder} \leftrightarrow \textit{l.elbow} \leftrightarrow \textit{l.wrist}\;, \\
    \mathcal{G}^{kin}_{r.arm}&:\; \textit{r.shoulder} \leftrightarrow \textit{r.elbow} \leftrightarrow \textit{r.wrist}\;, \\
    \mathcal{G}^{kin}_{l.leg}&:\; \textit{l.hip} \leftrightarrow \textit{l.knee} \leftrightarrow \textit{l.foot}\;, \\
    \mathcal{G}^{kin}_{r.leg}&:\; \textit{r.hip} \leftrightarrow \textit{r.knee} \leftrightarrow \textit{r.foot}\;.
\end{align}
Kinematic grammar focuses on connected body parts and works both forward and backward. Forward kinematics takes the last joint in a kinematic chain into account while backward kinematics reversely influences a joint in a kinematics chain from the next joint.

\textbf{Symmetry grammar} $\mathcal{G}^{sym}$ measure bilateral symmetry of human body (\textit{i.e.}, blue skeleton in Figure~\ref{fig:intro}), as human body can be divided into matching halves by drawing a line down the center; the left and right sides are mirror images of each other. 
\begin{align}
    \mathcal{G}^{sym}_{arm}&:\; \mathcal{G}^{kin}_{l.arm} \leftrightarrow \mathcal{G}^{kin}_{r.arm}\;, \\
    \mathcal{G}^{sym}_{leg}&:\; \mathcal{G}^{kin}_{l.leg} \leftrightarrow \mathcal{G}^{kin}_{r.leg}\;.
\end{align}

\textbf{Motor coordination grammar} $\mathcal{G}^{crd}$ represents movements of several limbs combined in a certain manner (\textit{i.e.}, green skeleton in Figure~\ref{fig:intro}). In this paper, we consider simplified motor coordination between human arm and leg. We define 2 coordination grammar to represent constraints on people coordinated movements:
\begin{align}
    \mathcal{G}^{crd}_{l\rightarrow r}&:\; \mathcal{G}^{kin}_{l.arm} \leftrightarrow \mathcal{G}^{kin}_{r.leg}\;, \\
    \mathcal{G}^{crd}_{r\rightarrow l}&:\; \mathcal{G}^{kin}_{r.arm} \leftrightarrow \mathcal{G}^{kin}_{l.leg}\;.
\end{align}

The RNN naturally supports chain-like structure, which provides a powerful tool for modeling our grammar formulations with deep learning. 
There are two states (forward/backfward directions) encoded in BRNN. At each time step $t$, with the input feature $a_t$, the output $y_t$ is determined by considering two-direction states $h^f_t$ and $h^b_t$:
\begin{equation}\small\begin{aligned}
    y_t = \phi(W^f_yh^f_t+ W^b_yh^b_t+b_y),
\end{aligned}\label{eq:BRNN}\end{equation}
where $\phi$ is the softmax function and the states $h^f_t, h^b_t$ are computed as:
\begin{equation}\small\begin{aligned}
    &h^f_t = \tanh(W^f_hh^f_{t-1}+ W^f_a a_t+b^f_h)\;,\\
    &h^b_t = \tanh(W^b_hh^b_{t+1}+ W^b_a a_t+b^b_h)\;,
\end{aligned}\end{equation}

As shown in Figure~\ref{fig:network}, we build a two-layer tree-like hierarchy of BRNNs for modeling our three grammar, where each of the BRNNs shares same equation in Eqn.\eqref{eq:BRNN} and the three grammar are represented by the edges between BRNNs nodes or implicitly encoded into BRNN architecture.

For the bottom layer, five BRNNs are built for modeling the five relations defined in kinematics grammar. More specifically, they accept the pose-aligned features from our base network as input, and generate estimation for a 3D joint at each time step. The information is forward/backfoward propagated efficiently over the two states with BRNN, thus the five Kinematics relations are implicitly modeled by the bi-directional chain structure of corresponding BRNN. Note that we take the advantages of recurrent natures of RNN for capturing our chain-like grammar, instead of using RNN for modeling the temporal dependency of sequential data.

For the top layer, totally four BRNN nodes are derived, two for symmetry relations and two for motor coordination dependencies. For the symmetry BRNN nodes, taking $\mathcal{G}^{sym}_{arm}$ node as an example, it takes the concatenated 3D-joints (totally 6 joints) from the $\mathcal{G}^{kin}_{l.arm}$ and $\mathcal{G}^{kin}_{r.arm}$ BRNNs in the bottom layer in all times as input, and produces estimations for the six 3D-joints taking their symmetry relations into account. Similarly, for the coordination nodes, such as $\mathcal{G}^{crd}_{l\rightarrow r}$, it leverages the estimations from $\mathcal{G}^{kin}_{l.arm}$ and $\mathcal{G}^{kin}_{r.leg}$ BRNNs and refines the 3D joints estimations according to coordination grammar.

In this way, we inject three kinds of human pose grammar into a tree-BRNN model and the final 3D human joints estimations are achieved by mean-pooling the results from all the nodes in the grammar hierarchy.

\section{Learning} \label{sec:learn}

\begin{figure}[ptb]
\centering
\includegraphics[width=\linewidth]{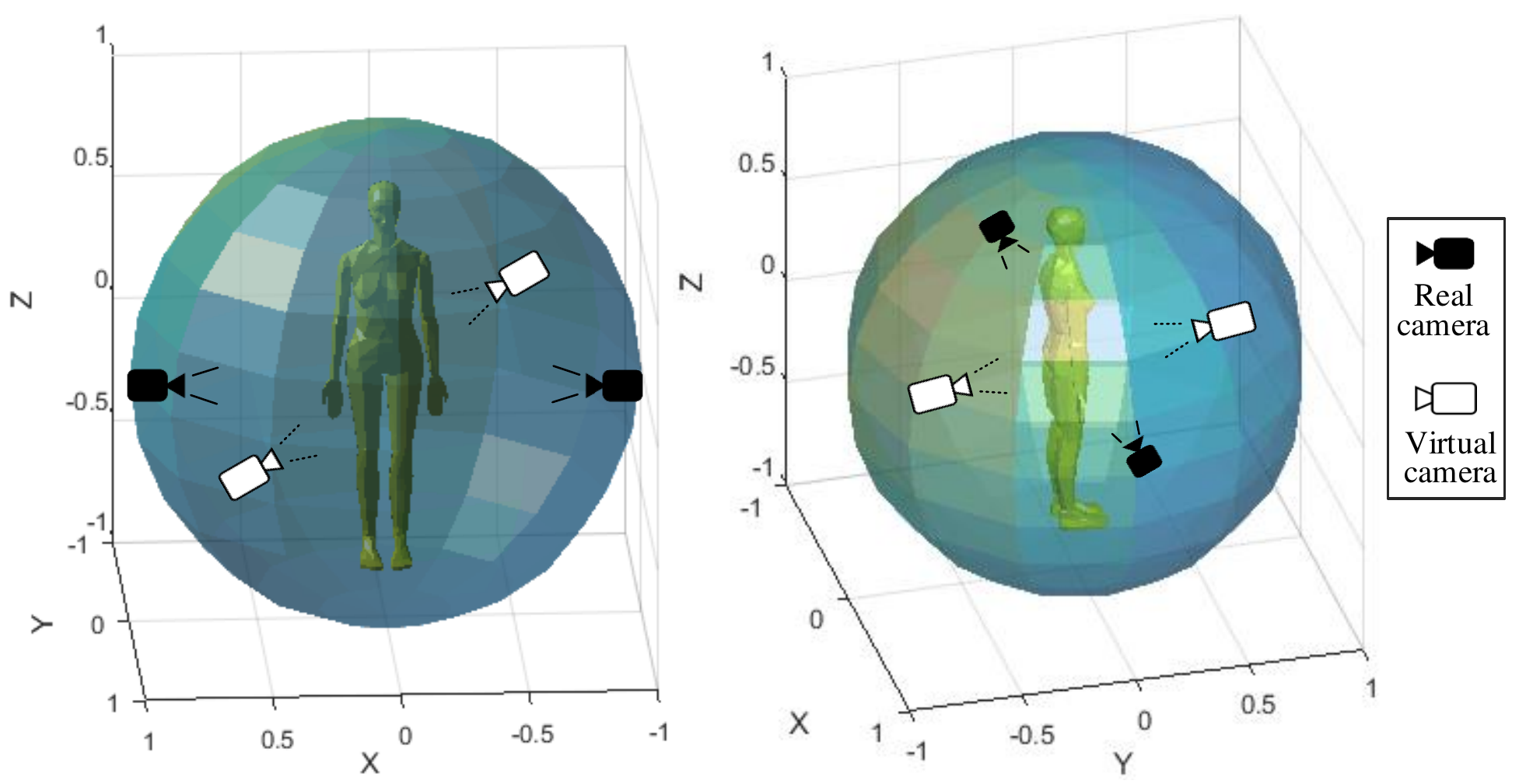}
\caption{Illustration of virtual camera simulation. The black camera icons stand for real camera settings while the white camera icons simulated virtual camera settings.}
\label{fig:camera}
\end{figure}

Given a training set $\Omega$:
\begin{equation}
    \Omega = \{(\hat{\mathbf{U}}^k, \hat{\mathbf{V}}^k):\,k = 1,\dots,N_\Omega\},
\end{equation}
where $\hat{\mathbf{U}}^k$ and $\hat{\mathbf{V}}^k$ denote ground-truth 2D and 3D pose pairs, we define the 2D-3D loss of learning the mapping function $f(\mathbf{U};\theta)$ as
\begin{equation} \begin{aligned}
    \theta^* &= \underset{\theta}{\arg\,\min}\; \ell(\Omega|\theta) \\
    &= \underset{\theta}{\arg\,\min}\;\sum_{k=1}^{N_\Omega} \|f(\hat{\mathbf{U}}^k;\theta) - \hat{\mathbf{V}}^k\|_2.
\end{aligned} \end{equation}
The loss measures the Euclidian distance between predicted 3D pose and true 3D pose.

The entire learning process consists of two steps: i) learning basic blocks in the base network with 2D-3D loss.
ii) attaching pose grammar network on the top of the trained base network, and fine-tune the whole network in an end-to-end manner.

\subsection{Pose Sample Simulator}

We conduct an empirical study on popular 3D pose estimation datasets (\textit{e.g.},
\textit{Human3.6M}, \textit{HumanEva}) and notice that there are usually limited number of cameras (4 on average) recording the human subject. This raises the doubt whether learning on such dataset can lead to a generalized 3D pose estimator applicable in other scenes with different camera positions. We believe that a data augmentation process will help improve the model performance and generalization ability. For this, we propose a novel Pose Sample Simulator (PSS) to generate additional training samples. The generation process consists of two steps: i) projecting ground-truth 3D pose $\hat{\mathbf{V}}$ onto virtual camera planes to obtain ground-truth 2D pose $\hat{\mathbf{U}}$, ii) simulating 2D pose detections $\mathbf{U}$ by sampling conditional probability distribution $p(\mathbf{U}|\hat{\mathbf{U}})$.

In the first step, we first specify a series of virtual camera calibrations. Namely, a virtual camera calibration is specified by quoting intrinsic parameters $K'$ from other real cameras and simulating reasonable extrinsic parameters (\textit{i.e.}, camera locations $T'$ and orientations $R'$). As illustrated in Figure~\ref{fig:camera}, two white virtual camera calibrations are determined by the other two real cameras. Given a specified virtual camera, we can perform a perspective projection of a ground-truth 3D pose $\hat{\mathbf{V}}$ onto the virtual camera plane and obtain the corresponding ground-truth 2D pose $\hat{\mathbf{U}}$.

\begin{figure}[ptb]
\centering
\includegraphics[width=\linewidth]{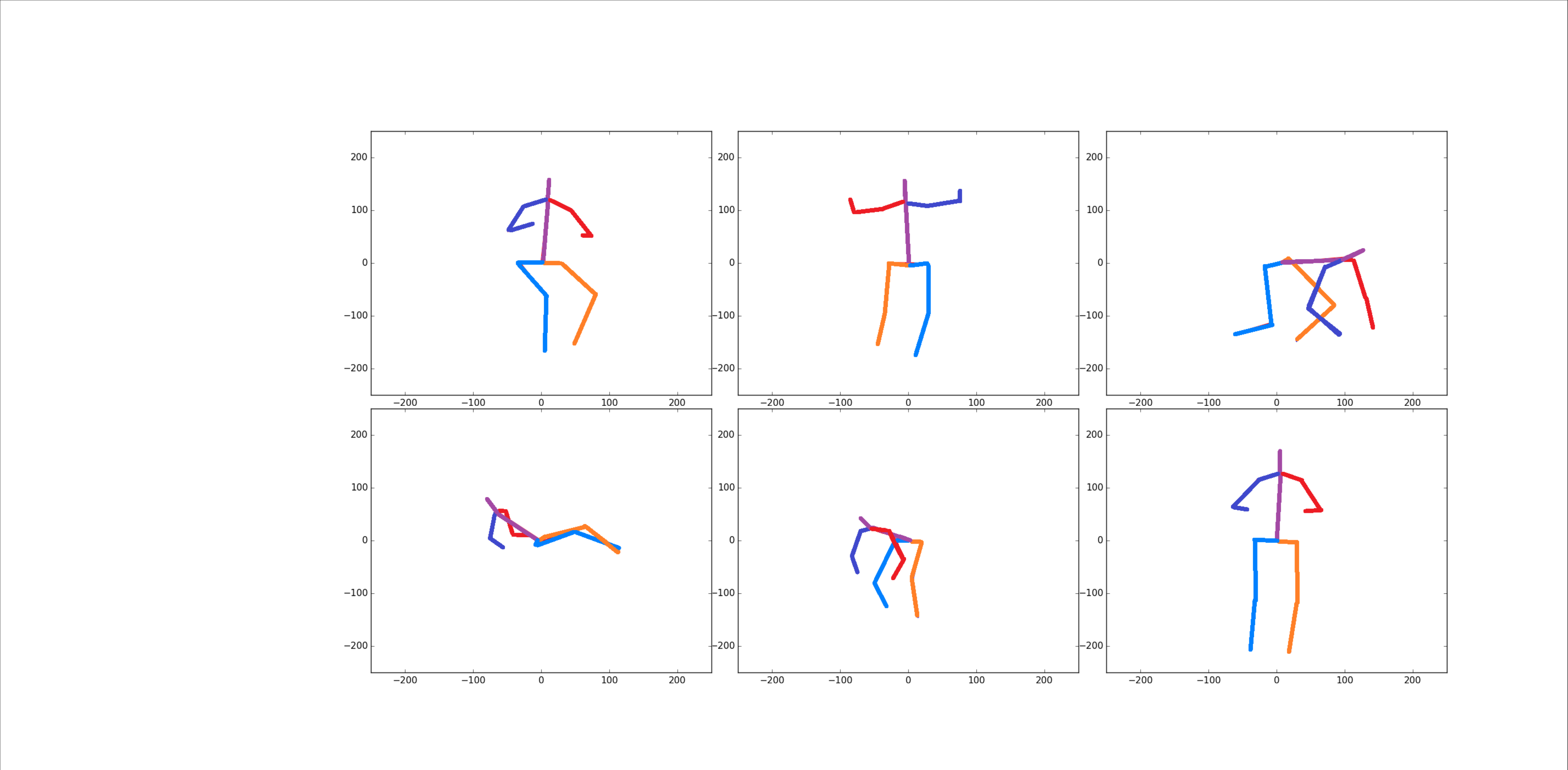}
\caption{Examples of learned 2D atomic poses in probability distribution $p(\mathbf{U}|\hat{\mathbf{U}})$.}
\label{fig:gmm}
\end{figure}

In the second step, we first model the conditional probability distribution $p(\mathbf{U}|\hat{\mathbf{U}})$ to mitigate the discrepancy between 2D pose detections $\mathbf{U}$ and 2D pose ground-truth $\hat{\mathbf{U}}$. Assuming $p(\mathbf{U}|\hat{\mathbf{U}})$ follows a mixture of Gaussian distribution, that is,
\begin{equation}
    p(\mathbf{U}|\hat{\mathbf{U}}) = p(\epsilon) = \sum_{j=1}^{N_G}\,\omega_{j}\, \mathbb{N}(\epsilon; \mu_j, \Sigma_j),
\end{equation}
where $\epsilon=\mathbf{U}-\hat{\mathbf{U}}$, $N_G$ denotes the number of Gaussian distributions, $\omega_{j}$ denotes a combination weight for the $j$-th component, $\mathbb{N}(\epsilon; \mu_j, \Sigma_j)$ denotes the $j$-th multivariate Gaussian distribution with mean $\mu_{j}$ and covariance $\Sigma_j$. As suggested in~\cite{andriluka20142d}, we set $N_G=42$. For efficiency issues, the covariance matrix $\Sigma_j$ is assumed to be in the form:
\begin{equation}
  \Sigma_j = \begin{bmatrix}
                \sigma_{j,1} &0 &0   \\
                0 & \ddots & 0       \\
                0 & 0 & \sigma_{j,i} \\
             \end{bmatrix},\;\;
  \sigma_{j,i} \in \mathbb{R}^{2\times2}
\end{equation}
where $\sigma_{j,i}$ is the covariance matrix for joint $u_i$ at $j$-th multivariate Gaussian distribution. This constraint enforces independence among each joint $u_i$ in 2D pose $\mathbf{U}$.

The probability distribution $p(\mathbf{U}|\hat{\mathbf{U}})$ can be efficiently learned using an EM algorithm, with E-step estimating combination weights $\omega$ and M-step updating Gaussian parameters $\mu$ and $\Sigma$. We utilizes K-means clustering to initialize parameters as a warm start. The learned mean $\mu_j$ of each Gaussian can be considered as an atomic pose representing a group of similar 2D poses. We visualize some atomic poses in Figure~\ref{fig:gmm}.

Given a 2D pose ground-truth $\hat{\mathbf{U}}$, we sample $p(\mathbf{U}|\hat{\mathbf{U}})$ to generate simulated detections $\mathbf{U}$ and thus use it augment the training set $\Omega$. By doing so we mitigate the discrepancy between the training data and the testing data. The effectiveness of our proposed PSS is validated in Section~\ref{sec:ablative}.

\section{Experiments} \label{sec:exp}

In this section, we first introduce datasets and settings for evaluation, and then report our results and comparisons with state-of-the-art methods, and finally conduct an ablation study on components in our method.

\begin{table*}[ptb]
\centering
\setlength{\tabcolsep}{4pt}
\resizebox{\textwidth}{!}{
\begin{tabular}{@{}lcccccccccccccccc@{}}
\toprule
\textbf{Protocol \#1} & Direct. & Discuss & Eating & Greet & Phone & Photo & Pose & Purch. & Sitting & SittingD. & Smoke & Wait & WalkD. & Walk & WalkT. & Avg.\\
\midrule
LinKDE (PAMI'16) & 132.7 & 183.6 & 132.3 & 164.4 & 162.1 & 205.9 & 150.6 & 171.3 & 151.6 & 243.0 & 162.1 & 170.7 & 177.1 & 96.6 & 127.9 & 162.1\\
Tekin~et al. (ICCV'16) & 102.4 & 147.2 & 88.8 & 125.3 & 118.0 & 182.7 & 112.4 & 129.2 & 138.9 & 224.9 & 118.4 & 138.8 & 126.3 & 55.1 & 65.8 & 125.0\\
Du~et al. (ECCV'16) & 85.1 & 112.7 & 104.9 & 122.1 & 139.1 & 135.9 & 105.9 & 166.2 & 117.5 & 226.9 & 120.0 & 117.7 & 137.4 & 99.3 & 106.5 & 126.5\\
Chen \& Ramanan (Arxiv'16) & 89.9 & 97.6 & 89.9 &  107.9 &  107.3 & 139.2 &  93.6 &  136.0 & 133.1 & 240.1 & 106.6 &  106.2 &  87.0 & 114.0 & 90.5 & 114.1 \\
Pavlakos~et al. (CVPR'17) & 67.4 & 71.9 & 66.7 & 69.1 & 72.0 & 77.0 & 65.0 & 68.3 & 83.7 & 96.5 & 71.7 & 65.8 & 74.9 & 59.1 & 63.2 & 71.9\\
Bruce~et al. (ICCV'17) & 90.1 & 88.2 & 85.7 & 95.6 & 103.9 & 92.4 & 90.4 & 117.9 & 136.4 & 98.5 & 103.0 & 94.4 & 86.0 & 90.6 & 89.5 & 97.5\\
Zhou~et al. (ICCV'17) & 54.8 & 60.7 & 58.2 & 71.4 & \bf{62.0} & \bf{65.5} & 53.8 & \bf{55.6} & 75.2 & 111.6 & 64.1 & 66.0 & \bf{51.4} & 63.2 & 55.3 & 64.9\\
Martinez~et al. (ICCV'17) & 51.8& 56.2& 58.1&	59.0& 69.5&	78.4& 55.2&	58.1&74.0&94.6&	62.3&59.1&65.1&49.5&52.4&62.9\\
\midrule
Ours & \bf{50.1}&  \bf{54.3}&	\bf{57.0}&	\bf{57.1}&	66.6& 73.3&	\bf{53.4}&	55.7&	\bf{72.8}&	\bf{88.6}&	\bf{60.3}&	\bf{57.7}&	62.7&	\bf{47.5}&	\bf{50.6}&	\bf{60.4}\\
\toprule
\textbf{Protocol \#2} & Direct. & Discuss & Eating & Greet & Phone & Photo & Pose & Purch. & Sitting & SittingD. & Smoke & Wait & WalkD. & Walk & WalkT. & Avg.\\
\midrule
Ramakrishna~et al.(ECCV'12) & 137.4 & 149.3 & 141.6 & 154.3 & 157.7 & 158.9 & 141.8 & 158.1 & 168.6 & 175.6 & 160.4 & 161.7 & 150.0 & 174.8 & 150.2 & 157.3\\
Bogo~et al.(ECCV'16) & 62.0 & 60.2 & 67.8 & 76.5 & 92.1 & 77.0 & 73.0 & 75.3 & 100.3 & 137.3 & 83.4 & 77.3 & 86.8 & 79.7 & 87.7 & 82.3\\
Moreno-Noguer (CVPR'17) & 66.1 & 61.7 & 84.5 & 73.7 & 65.2 & 67.2 & 60.9 & 67.3 & 103.5 & 74.6 & 92.6 & 69.6 & 71.5 & 78.0 & 73.2 & 74.0\\
Pavlakos~et al. (CVPR'17)  & -- & -- & -- & -- & -- & -- & -- & -- & -- & -- & -- & -- & -- & -- & -- & 51.9\\
Bruce~et al. (ICCV'17) & 62.8 & 69.2 & 79.6 & 78.8 & 80.8 & 72.5 & 73.9 & 96.1 & 106.9 & 88.0 & 86.9 & 70.7 & 71.9 & 76.5 & 73.2 & 79.5\\
Martinez~et al. (ICCV'17) & 39.5 & 43.2&46.4&	47.0&	51.0&	56.0&	41.4&	40.6&	56.5&	69.4&	49.2&	45.0&	49.5&	38.0&	43.1&	47.7\\
\midrule
Ours & \bf{38.2} & \bf{41.7}&	\bf{43.7}&	\bf{44.9}&	\bf{48.5}&	\bf{55.3}&	\bf{40.2}&	\bf{38.2}&	\bf{54.5}&	\bf{64.4}&	\bf{47.2}&	\bf{44.3}&	\bf{47.3}&	\bf{36.7}&	\bf{41.7}&	\bf{45.7}\\
\toprule
\textbf{Protocol \#3} & Direct. & Discuss & Eating & Greet & Phone & Photo & Pose & Purch. & Sitting & SitingD. & Smoke & Wait & WalkD. & Walk & WalkT. & Avg.\\
\midrule
Pavlakos~et al. (CVPR'17) & 79.2 & 85.2 & 78.3 & 89.9 & 86.3 & 87.9 & 75.8 & 81.8 & 106.4 & 137.6 & 86.2 & 92.3 & 72.9 & 82.3 & 77.5 & 88.6\\
Bruce~et al. (ICCV'17) & 103.9 & 103.6 & 101.1 & 111.0 & 118.6 & 105.2 & 105.1 & 133.5 & 150.9 & 113.5 & 117.7 & 108.1 & 100.3 & 103.8 & 104.4 & 112.1\\
Zhou~et al. (ICCV'17) & 61.4 & 70.7 & \bf{62.2} & 76.9 & \bf{71.0} & \bf{81.2} & 67.3 & 71.6 & 96.7 & 126.1 & \bf{68.1} & 76.7 & \bf{63.3} & 72.1 & 68.9 & 75.6\\
Martinez~et al. (ICCV'17) & 65.7 & 68.8 & 92.6 & 79.9 & 84.5 & 100.4 & 72.3 & 88.2 & 109.5 & 130.8 & 76.9 & 81.4 & 85.5 & 69.1 & 68.2 & 84.9\\
\midrule
Ours & \bf{57.5}&  \bf{57.8}& 81.6&	\bf{68.8}&	75.1&	85.8&	\bf{61.6}&	\bf{70.4}&	\bf{95.8}&	\bf{106.9}&	68.5&	\bf{70.4}&	73.8&	\bf{58.5}&	\bf{59.6}&	\bf{72.8}\\
\bottomrule
\end{tabular}
}
\caption{Quantitative comparisons of Average Euclidean Distance (mm) between the estimated pose and the ground-truth on \textit{Human3.6M} under \textit{Protocol \#1}, \textit{Protocol \#2} and \textit{Protocol \#3}. The best score is marked in \textbf{bold}.}
\label{tab:h36m}
\end{table*}

\subsection{Datasets}

We evaluate our method quantitatively and qualitatively on three popular 3D pose estimation datasets.

\textbf{Human3.6M}~\cite{ionescu2014human3} is the current largest dataset for human 3D pose estimation, which consists of 3.6 million 3D human poses and corresponding video frames recorded from 4 different cameras. Cameras are located at the front, back, left and right of the recorded subject, with around 5 meters away and 1.5 meter height. In this dataset, there are 11 actors in total and 15 different actions performed (\textit{e.g.}, greeting, eating and walking). The 3D pose ground-truth is captured by a motion capture (Mocap) system and all camera parameters (intrinsic and extrinsic parameters) are provided.

\textbf{HumanEva-I}~\cite{sigal2010humaneva} is another widely used dataset for human 3D pose estimation, which is also collected in a controlled indoor environment using a Mocap system. \textit{HumanEva-I} dataset has fewer subjects and actions, compared with \textit{Human3.6M} dataset.

\textbf{MPII}~\cite{andriluka20142d} is a challenging benchmark for 2D human pose estimation in the wild, containing a large amount of human images in the wild. We only validate our method on this dataset qualitatively since no 3D pose ground-truth is provided.

\subsection{Evaluation Protocols}

For \textbf{Human3.6M}, the standard protocol is using all 4 camera views in subjects S1, S5, S6, S7 and S8 for training and the same 4 camera views in subjects S9 and S11 for testing. This standard protocol is called \textit{protocol \#1}. In some works, the predictions are post-processed via a rigid transformation before comparing to the ground-truth, which is referred as \textit{protocol \#2}.

In above two protocols, the same $4$ camera views are both used for training and testing. This raise the question whether or not the learned estimator over-fits to training camera parameters. To validate the generalization ability of different models, we propose a new protocol based on different camera view partitions for training and testing. In our setting, subjects S1, S5, S6, S7, and S8 in 3 camera views are used for training while subjects S9 and S11 in the other camera view are selected for testing (down-sampled to 10fps). The suggested protocol guarantees that not only subjects but also camera views are different for training and testing, eliminating interferences of subject appearance and camera parameters, respectively. We refer our new protocol as \textit{protocol \#3}.

For \textbf{HumanEva-I}, we follow the previous protocol, evaluating on each action separately with all subjects. A rigid transformation is performed before computing the mean reconstruction error.

\subsection{Implementation Details}

We implement our method using Keras with Tensorflow as back-end. We first train our base network for 200 epoch. The learning rate is set as 0.001 with exponential decay and the batch size is set to 64 in the first step. Then we add the 3D-Pose Grammar Network on top of the base network and fine-tune the whole network together. The learning rate is set as $10^{-5}$ during the second step to guarantee model stability in the training phase. We adopt Adam optimizer for both steps.

We perform 2D pose detections using a state-of-the-art 2D pose estimator~\cite{newell2016stacked}. We fine-tuned the model on \textit{Human3.6M} and use the pre-trained model on \textit{HumanEva-I} and \textit{MPII}. Our deep grammar network is trained with 2D pose detections as inputs and 3D pose ground-truth as outputs. For \textit{protocol \#1} and \textit{protocol \#2}, the data augmentation is omitted due to little improvement and tripled training time. For \textit{protocol \#3}, in addition to the original 3 camera views, we further augment the training set with 6 virtual camera views on the same horizontal plane. Consider the circle which is centered at the human subject and locates all cameras is evenly segmented into 12 sectors with 30 degree angles each, and 4 cameras occupy 4 sectors. We generate training samples on 6 out of 8 unoccupied sectors and leave 2 closest to the testing camera unused to avoid overfitting. The 2D poses generated from virtual camera views are augmented by our PCSS. During each epoch, we will sample our learned distribution once and generate a new batch of synthesized data.

Empirically, one forward and backward pass takes 25 ms on a Titan X GPU and a forward pass takes 10 ms only, allowing us to train and test our network efficiently.
\begin{figure*}[htb]
 \begin{center}
 \includegraphics[width=\linewidth]{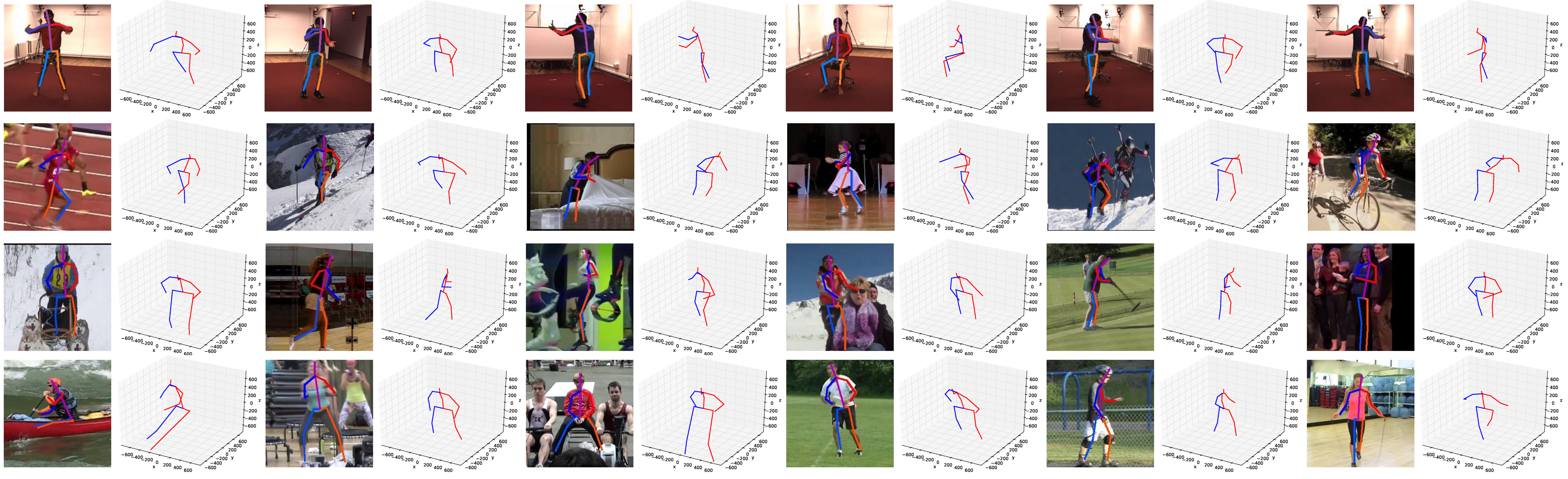}
 \caption{Quantitative results of our method on \textit{Human3.6M} and \textit{MPII}. We show the estimated 2D pose on the original image and the estimated 3D pose from a novel view. Results on \textit{Human3.6M} are drawn in the first row and results on \textit{MPII} are drawn in the second to fourth row. Best viewed in color.}
 \label{fig:result_table}
 \end{center}
\end{figure*}
\subsection{Results and Comparisons}

\begin{table}[ptb]
\begin{center}
\setlength{\tabcolsep}{4pt}
\resizebox{\linewidth}{!}{
\begin{tabular}{l|ccc|ccc|c}
\hline\thickhline
\multirow{2}{*}{Methods} & \multicolumn{3}{c|}{Walking} & \multicolumn{3}{c|}{Jogging} & \multirow{2}{*}{Avg.} \\
\cline{2-7}
& S1 & S2 & S3 & S1 & S2 & S3 & \\
\hline
Simo-Serra  et al. (CVPR'13) & 65.1 & 48.6 & 73.5 & 74.2 & 46.6 & 32.2 & 56.7\\
Kostrikov  et al. (BMVC'14) & 44.0 & 30.9 & 41.7 & 57.2 & 35.0 & 33.3 & 40.3 \\
Yasin  et al. (CVPR'16) & 35.8 & 32.4 & 41.6 & 46.6 & 41.4 & 35.4 & 38.9 \\
Moreno-Noguer (CVPR'17) &19.7& \textbf{13.0} &{\bf24.9} &39.7 &20.0 &21.0 &26.9\\
Pavlakos et al. (CVPR'17) & 22.3 & 19.5 & 29.7 & 28.9 & 21.9 & 23.8 & 24.3 \\
Martinez et al. (ICCV'17) & 19.7 & 17.4 & 46.8 & {\bf26.9} &18.2 & 18.6 & 24.6\\
\hline
Ours & {\bf19.4} & 16.8 & 37.4 & 30.4 &\bf{17.6} & \textbf{16.3} & \textbf{22.9}\\
\hline
\end{tabular}
}
\end{center}
\caption{Quantitative comparisons of the mean reconstruction error (mm) on \textit{HumanEva-I}. The best score is marked in \textbf{bold}.}
\label{table:humaneva}
\end{table}

\textbf{Human3.6M}. We evaluate our method under all three protocols. We compare our method with $10$ state-of-the-art methods~\cite{ionescu2014human3,tekin2016direct,du2016marker,chen20163d,sanzari2016bayesian,rogez2016mocap,bogo2016keep,pavlakos2017volumetric,Nie3DPoseCVPR17,zhou2017towards,martinez2017simple} and report quantitative comparisons in Table~\ref{tab:h36m}. From the results, our method obtains superior performance over the competing methods under all protocols.

To verify our claims, we re-train three previous methods, which obtain top performance under \textit{protocol \#1}, with \textit{protocol \#3}. The quantitative results are reported in Table.~\ref{tab:h36m}. The large drop of performance ($17\%$ -- $41\%$) of previous 2D-3D reconstruction models~\cite{pavlakos2017volumetric,Nie3DPoseCVPR17,zhou2017towards,martinez2017simple}, which demonstrates the blind spot of previous evaluation protocols and the over-fitting problem of those models.

Notably, our method greatly surpasses previous methods ($12 mm$ improvement over the second best under cross-view evaluation (\textit{i.e.}, \textit{protocol \#3}). Additionally, the large performance gap of~\cite{martinez2017simple} under \textit{protocol \#1} and \textit{protocol \#3} ($62.9 mm$ \textit{vs} $84.9 mm$) demonstrates that previous 2D-to-3D reconstruction networks easily over-fit to camera views. Our general improvements over different settings demonstrate our superior performance and good generalization.

\textbf{HumanEva-I}. We compare our method with $6$ state-of-the-art methods~\cite{simo2013joint,kostrikov2014depth,yasin2016dual,moreno20163d,pavlakos2017volumetric,martinez2017simple}. The quantitative comparisons on \textit{HumanEva-I} are reported in Table~\ref{table:humaneva}. As seen, our results outperforms previous methods across the vast majority of subjects and on average.

\textbf{MPII}. We visualize sampled results generated by our method on \textit{MPII} as well as \textit{Human3.6M} in Figure~\ref{fig:result_table}. As seen, our method is able to accurately predict 3D pose for both indoor and in-the-wild images.

\subsection{Ablation studies} \label{sec:ablative}

\begin{table}[ptb]
\centering
\resizebox{\linewidth}{!}{
\begin{tabular}{c|l|c|c}
\hline\thickhline
\multirow{2}{*}{Component} & Variants & Error (mm) & $\Delta$\\
\cline{2-4}
& Ours, full & 72.8 & --\\
\hline
\multirow{3}{*}{Pose grammar}&w/o. grammar & 75.1 & 2.3\\
&w. kinematics & 73.9 & 1.1\\
&w. kinematics+symmetry & 73.2 & 0.4\\
\hline
\multirow{3}{*}{PSS}&w/o. extra 2D-3D pairs & 82.6 & 9.8\\
&w. extra 2D-3D pairs, GT & 76.7 & 3.9\\
&w. extra 2D-3D pairs, simple  & 78.0 & 5.2\\
\hline
\multirow{4}{*}{PSS Generalization} & Bruce~et al. (ICCV'17) w/o. & 112.1 & --\\
& Bruce~et al. (ICCV'17) w. & 96.3 & 15.8\\
& Martinez et al. (ICCV'17) w/o. & 84.9 & --\\
& Martinez et al. (ICCV'17) w. & 76.0 & 8.9\\
\hline
\end{tabular}
}
\caption{Ablation studies on different components in our method. The evaluation is performed on \textit{Human3.6M} under \textit{Protocol \#3}. See text for detailed explanations.}
\label{tab:ablative}
\end{table}

We study different components of our model on \textit{Human 3.6M} dataset under \textit{protocol \#3}, as reported in Table~\ref{tab:ablative}.

\textbf{Pose grammar}. We first study the effectiveness of our grammar model, which encodes high-level grammar constraints into our network. First, we exam the performance of our baseline by removing all three grammar from our model, the error is $75.1 mm$. Adding the kinematics grammar provides parent-child relations to body joints, reducing the error by $1.6\%$ ($75.1 mm\rightarrow 73.9 mm$). Adding on top the symmetry grammar can obtain an extra error drops ($73.9 mm\rightarrow 73.2 mm$). After combing all three grammar together, we can reach an final error of $72.8 mm$.

\textbf{Pose Sample Simulator} (PSS). Next we evaluate the influence of our 2D-pose samples simulator. Comparing the results of only using the data from original 3 camera views in \textit{Human 3.6M} and the results of adding samples by generating ground-truth 2D-3D pairs from 6 extra camera views, we see an $7\%$ errors drop ($82.6 mm\rightarrow76.7 mm$), showing that extra training data indeed expand the generalization ability. Next, we compare our Pose Sample Simulator to a simple baseline, \textit{i.e.}, generating samples by adding random noises to each joint, say an arbitrary Gaussian distribution or a white noise. Unsurprisingly, we observe a drop of performance, which is even worse than using the ground-truth 2D pose. This suggests that the conditional distribution $p(E|\hat{E})$ helps bridge the gap between detection results and ground-truth. Furthermore, we re-train models proposed in~\cite{Nie3DPoseCVPR17,martinez2017simple} to validate the generalization of our PSS. Results also show a performance boost for their methods, which confirms the proposed PSS is a generalized technique. Therefore, this ablative study validates the generalization as well as effectiveness of our PSS.

\section{Conclusion}

In this paper, we propose a pose grammar model to encode the mapping function of human pose from 2D to 3D. Our method obtains superior performance over other state-of-the-art methods by explicitly encoding human body configuration with pose grammar and a generalized data argumentation technique. We will explore more interpretable and effective network architectures in the future.

{
\small
\bibliographystyle{aaai}
\bibliography{3dpose}
}

\end{document}